\documentclass{article}

\usepackage{microtype}
\usepackage{graphicx}
\usepackage{subcaption}
\usepackage{booktabs} 

\usepackage{hyperref}
\usepackage{url}

\usepackage[accepted]{icml2026}

\usepackage{amsmath}
\usepackage{amssymb}
\usepackage{mathtools}
\usepackage{amsthm}

\usepackage{booktabs}
\usepackage{multirow}
\usepackage{colortbl}
\usepackage{xcolor}
\usepackage{graphicx}

\definecolor{darkgray}{gray}{0.85}   
\definecolor{lightgray}{gray}{0.95}   

\usepackage[capitalize,noabbrev]{cleveref}

\theoremstyle{plain}

\theoremstyle{definition}

\theoremstyle{remark}

\usepackage[textsize=tiny]{todonotes}

\icmltitlerunning{HyLRA: Hybrid Layer Reuse Attention for Efficient Long-Context Inference}

\begin{document}

\twocolumn[ \icmltitle{HyLRA: Hybrid Layer Reuse Attention for Efficient \\ Long-Context Inference}



\icmlsetsymbol{equal}{*}
\icmlsetsymbol{intern}{$\dagger$}

\begin{icmlauthorlist} 
\icmlauthor{Xuan Ai}{intern,equal,univ} 
\icmlauthor{Qingqing Yang}{equal,comp} 
\icmlauthor{Peng Wang}{comp} 
\icmlauthor{Lei Deng}{comp} 
\icmlauthor{Lin Zhang}{comp} 
\icmlauthor{Renhai Chen}{comp} 
\icmlauthor{Gong Zhang}{comp} 
\end{icmlauthorlist}

\icmlaffiliation{univ}{City University of Hong Kong} 
\icmlaffiliation{comp}{Theory Lab, Huawei Technologies}

\icmlcorrespondingauthor{Xuan Ai}{xuanai2-c@my.cityu.edu.hk}


\vskip 0.3in ]

\printAffiliationsAndNotice{\icmlEqualContribution\textsuperscript{$\dagger$}Work done during Xuan Ai's research internship at Huawei }




\begin{abstract} 
Long-context inference in Large Language Models (LLMs) is bottlenecked by the quadratic computation complexity of attention and the substantial memory footprint of Key-Value (KV) caches. While existing sparse attention mechanisms attempt to mitigate this by exploiting inherent sparsity, they often rely on rigid patterns or aggressive pruning, failing to achieve an optimal balance between efficiency and accuracy. In this paper, we introduce {\bf HyLRA} ({\bf Hy}brid {\bf L}ayer {\bf R}euse {\bf A}ttention), a novel framework driven by layer-wise sparsity profiling. Our empirical analysis uncovers a dual characteristic in attention mechanics: \textit{intra-layer sensitivity}, where specific layers necessitate full attention to prevent feature distortion, and \textit{inter-layer similarity}, where consecutive layers share substantial critical tokens. Based on these observations, HyLRA employs an offline dynamic programming approach to derive an optimal layer-wise policy. This hybrid strategy retains full attention for sensitive layers to ensure robustness, while enabling tolerant layers to bypass quadratic calculations by directly reusing top-$k$ indices from preceding layers. This approach allows LLMs to restrict computation to the most critical tokens, effectively overcoming the quadratic bottleneck of dense attention. Extensive evaluations demonstrate that HyLRA improves inference throughput by 6\%--46\% while maintaining comparable performance (with $<1\%$ accuracy degradation), consistently outperforming state-of-the-art sparse attention methods. HyLRA is open source at \href{https://anonymous.4open.science/r/unified-cache-management-CF80/}{\texttt{/r/unified-cache-management-CF80/}}.

\end{abstract}

\section{Introduction}

Large Language Models (LLMs) have fundamentally revolutionized natural language processing, exhibiting remarkable capabilities in complex reasoning, in-context learning, and long-form generation \citep{openai2024gpt4o, geminiteam2025gemini3}. Fueled by the growing demand for processing massive information, the context windows of LLMs have expanded dramatically—from 4K to over 1 million tokens \citep{DBLP:conf/iclr/0055YZA25}. While such extended context empowers applications like long-document summarization and repository-level code understanding, it imposes a severe bottleneck during the decoding phase. As the sequence length increases, the auto-regressive nature of LLMs suffers from quadratic computational complexity in the attention mechanism and linear memory scaling driven by the expanding KV cache \citep{QUEST, HATA}.

To mitigate these overheads, sparse attention mechanisms have emerged as a promising solution. By selectively processing only a small fraction of tokens with high attention scores (i.e., the top-$k$), these methods theoretically preserve generation quality while reducing computation \citep{ge2023model, DBLP:conf/emnlp/GuptaDGCB21}. However, existing approaches often face a dilemma between efficiency and accuracy. Many methods identify critical tokens independently at each layer, incurring cumulative runtime overhead due to repetitive importance scoring and sorting. This per-layer overhead often diminishes the practical latency reduction, counteracting the benefits of sparsity. Others rely on rigid patterns, such as fixed sliding windows \citep{StreamingLLM}, or employ aggressive eviction policies \citep{H2O}, both of which risk discarding critical information and degrading performance on complex tasks. Critically, most prior works remain layer-agnostic, applying uniform sparsification strategies across all layers and failing to exploit the inherent structural redundancy at different model depths.

In this paper, we propose \textbf{Hy}brid \textbf{L}ayer \textbf{R}euse \textbf{A}ttention ({\bf HyLRA}), a framework grounded in a novel perspective on layer-wise attention interpretability. Through extensive offline profiling, we observe a distinct dichotomy in attention patterns. We identify \textit{sensitive layers}, which exhibit high susceptibility to approximation errors; for these layers, aggressive sparsification leads to significant feature distortion, necessitating full attention to preserve global context. In contrast, \textit{tolerant layers} demonstrate high robustness and, crucially, strong inter-layer similarity. The critical tokens in these layers overlap significantly with those of preceding layers, indicating that attention patterns are locally stable and can be efficiently reused. This insight reveals that a uniform sparse policy is suboptimal: sensitive layers demand full precision, whereas tolerant layers can be efficiently approximated via index reuse with negligible performance degradation.

Leveraging this insight, HyLRA accelerates long-context inference via a profile-guided hybrid strategy. We formulate the layer-wise configuration as a path-finding problem on a similarity matrix, solving it via Dynamic Programming (DP) to derive a globally optimal policy. During inference, HyLRA executes this hybrid policy: it selectively performs standard full attention on sensitive layers to ensure robustness, and for tolerant layers, it bypasses quadratic calculations by directly reusing the top-$k$ indices from the preceding layer. This approach yields a dual advantage: it eliminates the computational overhead of redundant token selection while ensuring generation quality by maintaining full-precision attention where it matters most.

To evaluate the effectiveness of HyLRA, we conducted extensive experiments across diverse LLM models and long-context benchmarks. Results demonstrate that HyLRA significantly outperforms state-of-the-art sparse attention baselines, achieving up to \textbf{1.45$\times$} speedup with negligible performance degradation.

In summary, our contributions are as follows:

\begin{itemize}
\item We characterize the layer-wise heterogeneity in LLMs, identifying a clear distinction between sensitive layers (prone to feature distortion) and tolerant layers (exhibiting high inter-layer similarity).

\item We propose HyLRA, a hybrid attention framework that optimizes inference by retaining full attention for sensitive layers while employing an efficient top-$k$ index reuse mechanism for tolerant layers.

\item Extensive experiments demonstrate that HyLRA achieves a superior Pareto frontier between efficiency and accuracy compared to existing methods, enabling efficient inference for long-context LLMs.
\end{itemize}

\section{Related Work}

\subsection{Long-context LLM Inference}

Driven by the growing demand for extensive information processing, scaling the context window has become a focal point for modern LLMs, empowering applications such as repository-level code understanding, long-document summarization, and complex multi-turn dialogue \citep{bai2023longbench, chen2023longlora}.
State-of-the-art commercial models, including GPT-4o, Gemini~3, and Claude~4.5, now support ultra-long contexts ranging from hundreds of thousands to millions of tokens \citep{openai2024gpt4o, geminiteam2025gemini3, anthropic2025claude45}.
However, deploying these capabilities in production environments presents significant challenges. The computational and memory overhead of long-context inference scales aggressively with sequence length, primarily due to the expanding Key-Value (KV) cache \citep{kwon2023efficient}.
Specifically, the linear growth of memory requirements creates a substantial bottleneck for system throughput and latency \citep{dao2022flashattention}. To address this, HyLRA aims to accelerate long-context inference by effectively exploiting the sparsity within the KV cache.

\subsection{Sparse Attention}

Sparse attention mechanisms exploit the inherent redundancy in attention distributions by processing only a fraction of relevant tokens rather than scanning the entire KV cache. Despite their promise, balancing efficiency and accuracy remains a persistent challenge. 
One category of methods focuses on cache eviction. StreamingLLM \citep{StreamingLLM} utilizes attention sinks and a sliding window to handle infinite-length streams, while H2O \citep{H2O} retains only tokens with high accumulated attention scores. Similarly, PyramidInfer \citep{PyramidInfer} constructs a pyramidal cache by progressively pruning KV pairs in deeper layers. However, by permanently discarding tokens based on heuristics, these methods risk evicting information essential for future generation, potentially causing irreversible context loss.

To mitigate this risk, other approaches employ dynamic token selection. RetrievalAttention \citep{RetrievalAttention} and ClusterKV \cite{ClusterKV} utilize vector search and semantic clustering, respectively, to retrieve critical tokens on the fly. Quest \citep{QUEST} estimates token importance at the block level using min-max statistics, while InfiniGen \citep{InfiniGen} speculatively pre-fetches cache blocks by leveraging input similarity between consecutive layers. TidalDecode \citep{TidalDecode} identifies relevant tokens via full attention at specific selection layers and reuses them subsequently. ShadowKV \citep{DBLP:conf/icml/SunCB0Z0DCC25} maintains low-rank compressed keys to support larger batch sizes. 
Crucially, however, most existing strategies overlook the structural heterogeneity of attention across different depths, often applying uniform selection policies that fail to capture layer-wise variations. Addressing this limitation, we propose HyLRA, which introduces a hybrid inter-layer top-$k$ reuse mechanism to efficiently accelerate long-context attention while retaining the full KV cache to ensure accuracy.

\section{Methodlogy}

In this section, we first motivate the design of HyLRA through an offline profiling analysis of attention patterns. We then detail the proposed framework, which employs dynamic programming to derive a hybrid inter-layer top-$k$ reuse mechanism. Finally, we demonstrate how HyLRA extends to block-level sparsity to optimize hardware efficiency.

\subsection{Intra-Layer Sensitivity}

\begin{figure}[t] 
    \centering
    \begin{subfigure}[b]{0.49\linewidth} 
        \centering
        \includegraphics[width=\linewidth]{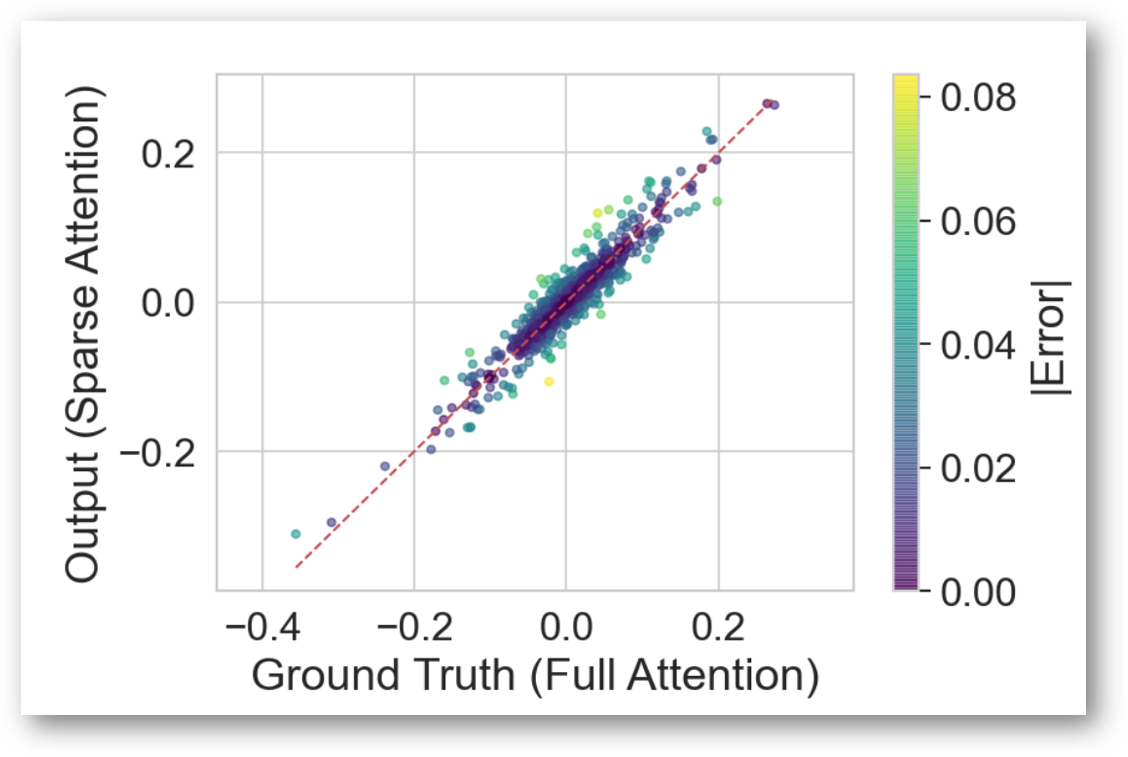} 
        \caption{Sensitivity of layer 0} 
        \label{fig00:sub_a}
    \end{subfigure}
    \hfill 
    \begin{subfigure}[b]{0.465\linewidth}
        \centering
        \includegraphics[width=\linewidth]{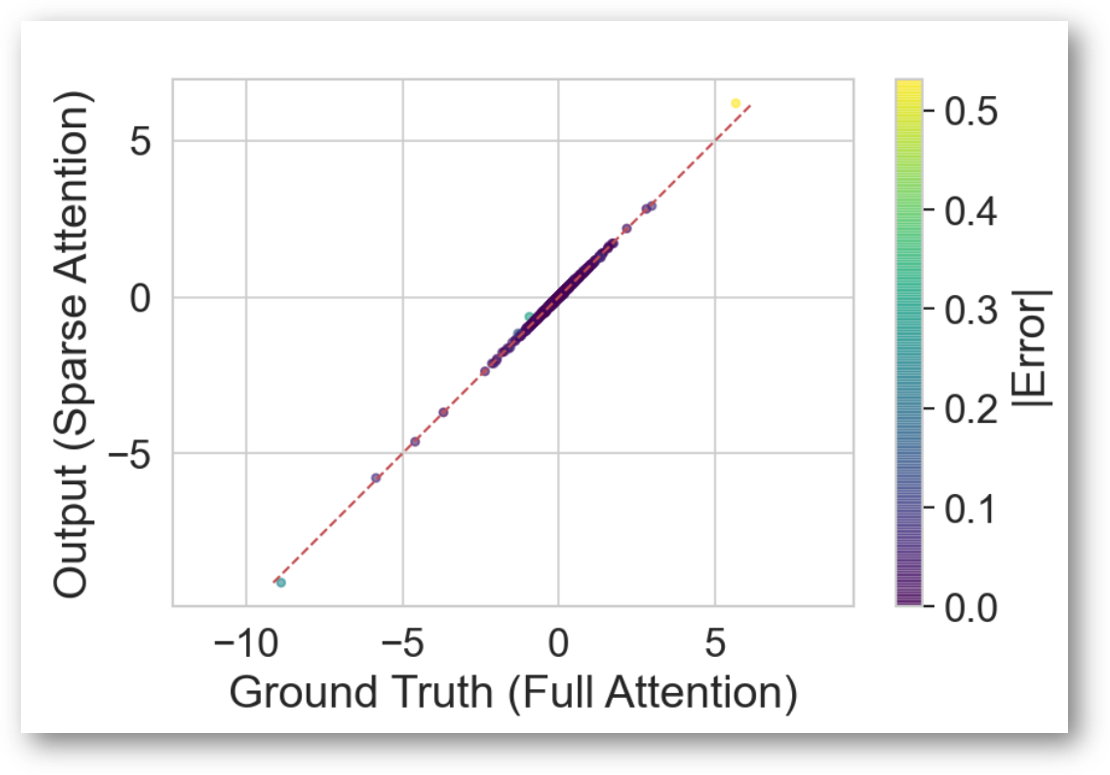}
        \caption{Sensitivity of layer 7}
        \label{fig00:sub_b}
    \end{subfigure}
    
    \caption{Sensitivity analysis of each layer in Qwen3-32B. We measure the sensitivity of the $l$-th layer by applying Top-$k$ (k = 2,048 tokens) sparse attention to it, while maintaining full attention for all preceding layers ($0$ to $l-1$). The impact is evaluated by measuring the deviation in the input to layer $l+1$. (a) Layer 0 exhibits significant sensitivity (high KL divergence with RNMSE = 0.25). Layer 7 demonstrates low sensitivity (low KL divergence with RNMSE = 0.016).}
    \label{fig00}
\end{figure}

LLMs are built upon multiple transformer layers, and the core of each layer is the attention module, which is computed as
\begin{align}\label{attention}
Attention  = \text{Softmax}(\frac{Q\cdot K^T}{\sqrt{d}}) V.
\end{align}
However, this operation becomes the primary bottleneck during long-context inference. For instance, it consumes up to 70\% of total runtime when the sequence length reaches 32K \citep{ribar2024sparq}, due to both quadratic computational complexity and memory bandwidth constraints \citep{ribar2024sparq}.
To address this, top-$k$ attention \citep{gupta-etal-2021-memory} exploits the inherent sparsity of attention distributions to skip computations for low-scoring tokens.

We observe that a uniform application of top-$k$ attention is suboptimal, as the \textit{sensitivity} to approximation errors varies significantly across layers. As illustrated in Figure~\ref{fig00}, we categorize layers into two distinct groups based on how their output features are affected by sparsification.

\textbf{Sensitive layers.} These layers exhibit high susceptibility to approximation errors. As shown in Figure~\ref{fig00}(a) (e.g., Layer~0), applying top-$k$ sparsification leads to a significant distortion in the output features (RNMSE = 0.25) and a large distributional shift. Aggressive sparsification here risks disrupting the foundational feature extraction, leading to noticeable error propagation to subsequent layers. Consequently, sensitive layers necessitate full attention to preserve critical information.

\textbf{Tolerant layers.} These layers demonstrate high robustness to sparsification. As depicted in Figure~\ref{fig00}(b) (e.g., Layer~7), restricting attention to the top-$k$ tokens introduces negligible deviation from the ground truth (RNMSE = 0.016). For these layers, the sparse approximation maintains high fidelity to the original full attention output, allowing for substantial reductions in computational and memory overhead without compromising accuracy.

This sensitivity heterogeneity motivates our layer-adaptive strategy, which selectively applies full attention to sensitive layers and sparse attention to tolerant layers, achieving an effective trade-off between efficiency and accuracy.

\subsection{Inter-Layer Similarity}

\begin{figure}[t]
    \centering
    \includegraphics[width=0.8\linewidth]{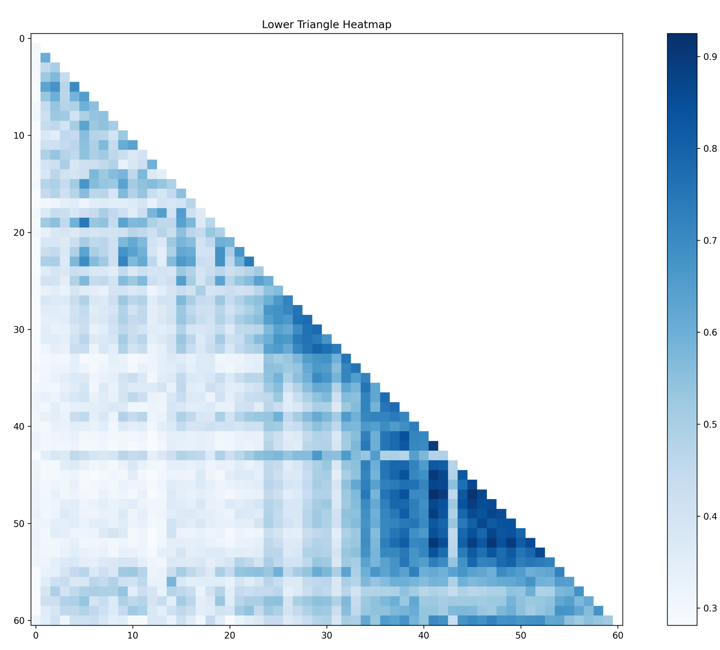}
    
    \caption{Visualization of layer-wise attention similarity. We calculate the overlap ratio of the top-$k$ ($k$ = 2,048) tokens with the highest attention scores between any pair of layers in DeepSeek-R1. The heatmap shows the pairwise overlap ratio, where the x-axis and y-axis represent layer indices. The dark regions along the diagonal indicate that consecutive layers tend to share a large number of critical tokens, suggesting a high degree of local consistency in attention distributions.}
    \label{fig01} 
\end{figure}

Beyond intra-layer sparsity, another key observation motivating HyLRA is the strong inter-layer similarity in attention patterns across adjacent layers. 
To quantify this observation, we conduct a needle-in-the-haystack test using DeepSeek-R1 \citep{deepseek2025} on the PG-19-mini \citep{rae2019compressive} with a context length of 60K tokens. We perform standard full attention across all layers and identify the set of top-$k$ (where $k=2,048$) tokens with the highest attention scores for each layer. Let $S_i$ and $S_j$ denote the sets of top-$k$ token indices in layer $i$ and layer $j$, respectively. The overlap ratio is defined as
\begin{align}
Overlap_{i,j} = \frac{|S_i \cap S_j|}{k}.
\end{align}
Figure~\ref{fig01} shows the overlap ratios for all pairs of transformer layers, where a higher overlap ratio indicates that most of the top-$k$ tokens are identical in the two layers. We find that the top-$k$ critical tokens in a given layer frequently exhibit substantial overlap with those of the preceding layer, particularly in deeper layers. Such inter-layer consistency suggests that independently identifying top-$k$ selections at every layer may be redundant.

Motivated by this observation, we propose reusing the top-$k$ indices from the preceding layer for layers that exhibit high inter-layer similarity. When the attention distribution of the current layer aligns sufficiently with that of its predecessor, directly inheriting the top-$k$ indices introduces negligible approximation error while eliminating the computational overhead of redundant token selection.

This inter-layer reuse strategy complements intra-layer sparsity by exploiting structural coherence across layers, thereby further accelerating attention computation without compromising accuracy.

\subsection{Offline Optimal Policy via Dynamic Programming}

\begin{figure}[t]
    \centering
    \includegraphics[width=0.68\linewidth]{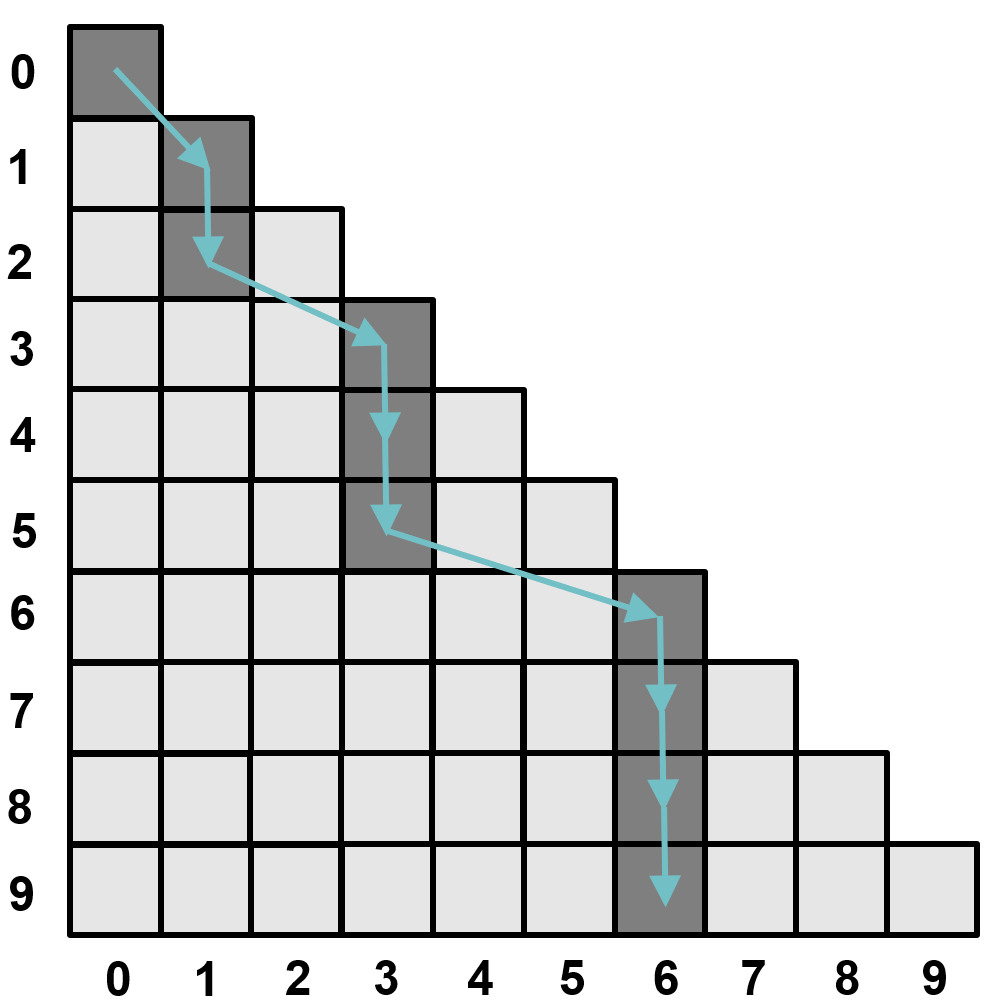}
    
    \caption{Visualization of offline policy optimization using a toy example with 10 layers. The grid maps the current layer (y-axis) against the reuse source (x-axis). The cyan path traces the optimal policy: diagonal steps to dark gray cells (e.g., layers 0, 3, 6) denote "Reset Jumps" (full attention refresh), while vertical steps to light gray cells denote "Index Reuse" (inheriting indices). This trajectory globally minimizes full attention frequency while maximizing cumulative similarity to ensure accuracy.}
    \label{fig02} 
\end{figure}

\begin{algorithm}[t]
\caption{Offline Policy Optimization via Dynamic Programming}
\label{alg00}
\begin{algorithmic}[1]
\REQUIRE{Similarity matrix $\mathcal{M}$, threshold $\theta$, number of layers $L$}

\FOR{all $j \in [0, L-1)$}
\FOR{each reachable state $(i,j)$ at layer $j$}
\STATE \textcolor{blue!60}{$\triangleright$ Vertical Move: Index Reuse}
\IF{$\mathcal{M}[i][j+1] \ge \theta$}
\STATE Insert candidate $(C[i][j], S[i][j] + \mathcal{M}[i][j+1])$ into $dp[i][j+1]$
\ENDIF
\STATE \textcolor{blue!60}{$\triangleright$ Reset Jump: Full Attention}
\STATE Insert candidate $(C[i][j] + 1, S[i][j] + 1)$ into $dp[j+1][j+1]$
\ENDFOR
\ENDFOR
\STATE Select optimal state $(i^*, L-1)$ minimizing $C$ and maximizing $S$
\STATE Find the optimal path $P^*$ and the corresponding policy $\pi$
\end{algorithmic}
\end{algorithm}

To determine the optimal layer-wise configuration policy $\pi$, we formulate the decision process as a path-finding problem over a similarity matrix derived from offline profiling, which we solve via dynamic programming.

\textbf{Similarity Matrix Construction.} 

We begin by constructing a lower-triangular similarity matrix $\mathcal{M} \in \mathbb{R}^{L \times L}$ based on profiling data. The horizontal axis represents the source layers, while the vertical axis denotes the target layers. Each entry $\mathcal{M}[i][j]$ (where $0\leq i\leq j <L$) quantifies the overlap ratio between the top-$k$ indices of the current layer $j$ and those of a preceding layer $i$. 
Diagonal entries ($i=j$) are set to 1, representing full attention, whereas off-diagonal entries reflect the fidelity of reusing top-$k$ indices across layers. This lower-triangular structure effectively defines the search space for optimizing the layer-wise reuse policy.

\textbf{Graph Traversal Formulation.}

We model each candidate policy as a valid path $P$ traversing the matrix $\mathcal{M}$, originating from the state (0, 0) and terminating at the final transformer layer. A state $(i,j)$ with $0\leq i < j <L$ on the path implies that layer $j$ reuses the top-$k$ indices from layer $i$, while a diagonal state $(j,j)$ corresponds to executing full attention at layer $j$.

Each path is characterized by two metrics: $C(P)$, the total count of full-attention computations, and $S(P)$, the cumulative similarity score, defined as $S(P) = \sum_{(i,j)\in P}\mathcal{M}[i][j]$. To guarantee performance, we impose a strict constraint that every state on the path must satisfy a similarity threshold $\mathcal{M}[i][j] \ge \theta$.

Our objective is to identify a path $P$ that yields the optimal trade-off between efficiency and fidelity. Specifically, we aim to minimize the computational cost while maximizing the cumulative similarity among all minimal-cost paths. This is formulated as the following constrained optimization problem:
\begin{align}
& C^* = \min_{P} C(P), \\
& P^* = \arg\max_{P: C(P)=C^*} S(P), \\
& \text{s.t.} \quad \mathcal{M}[i][j] \ge \theta,\ \forall (i,j)\in P.
\end{align}
This formulation explicitly prioritizes the reduction of full-attention operations while preserving accuracy through similarity-constrained reuse, naturally lending itself to an efficient dynamic programming solution.

From any given state $(i,j)$, the path can evolve via one of two transitions:

\textbf{Vertical Move (Index Reuse).} A vertical transition from state $(i,j)$ to $(i,j+1)$ indicates that layer $j+1$ inherits the top-$k$ indices from layer $i$. This transition is valid only if the similarity score satisfies $\mathcal{M}[i][j+1]\geq \theta$, where $\theta$ is the predefined fidelity threshold. Such a move incurs no additional full-attention cost and adds $\mathcal{M}[i][j+1]$ to the cumulative similarity score. The DP update is formalized as
\begin{align}
&candidate \gets (C[i][j], S[i][j] + \mathcal{M}[i][j+1]), \\
&dp[i][j+1] \gets dp[i][j+1] \cup \{candidate\}.
\end{align}

\textbf{Reset Jump (Full Attention).}
A reset jump from $(i,j)$ to the diagonal state $(j+1,j+1)$ signifies executing full attention at layer $j+1$ to refresh the top-$k$ indices. This action resets the reuse chain and increments the full-attention count by one. Accordingly, it contributes a similarity value of $\mathcal{M}[j+1][j+1]=1$, reflecting the exactness of full attention. The corresponding DP update is:
\begin{align}
&candidate \gets (C[i][j] + 1, S[i][j] + 1), \\
&dp[j+1][j+1] \gets dp[j+1][j+1] \cup \{candidate\}.
\end{align}

\textbf{Dynamic Programming Solution.}

By exploiting the layered structure of the similarity matrix, we solve the optimization problem sequentially. The algorithm propagates all reachable states from layer $j$ to layer $j+1$ using the defined transitions.

To efficiently enforce the objective, we maintain only the lexicographically optimal path for each state. During updates, any candidate path arriving at $(i,j)$ is compared against the existing best solution; candidates are discarded if they are dominated — possessing a higher full-attention count or an equal count with lower cumulative similarity. This strict pruning ensures that the propagated state always represents the path that primarily minimizes cost and secondarily maximizes fidelity.

Upon processing all layers, we select the optimal path from the final layer, directly yielding the optimal layer-wise execution policy $\pi$. Figure~\ref{fig02} shows the offline policy optimization using a toy example with 10 layers.

\subsection{Hybrid Inter-Layer Top-$K$ Reuse Attention}

\begin{algorithm}[t]
\caption{Hybrid Inter-Layer Top-$K$ Reuse Attention} 
\label{alg01}
\begin{algorithmic}[1]
\REQUIRE{Query $Q \in \mathbb{R}^{1 \times d}$ at current decoding step, key cache $K^{cache}\subset \mathbb{R}^{N \times d}$, value cache $V^{cache}\subset \mathbb{R}^{N \times d}$, layer-wise policy $\pi = \{a_1, \dots, a_L\}$, and top-$k$ budget $B$}
\FOR{each transformer layer $l$}
\IF{$a_l = \text{Full}$}
\STATE \textcolor{blue!60}{$\triangleright$ Calculate qk score}
\STATE $S \gets \operatorname{Score}(Q, K^{cache})$
\STATE \textcolor{blue!60}{$\triangleright$ Obtain top-$k$ indices}
\STATE $\mathcal{I}_l \gets \operatorname{TopK} (S,B)$
\STATE \textcolor{blue!60}{$\triangleright$ Calculate standard full attention output}
\STATE $O \leftarrow \operatorname{Softmax}(S)\cdot V^{cache}$
\ENDIF
\IF{$a_l = \text{Reuse}$}
\STATE \textcolor{blue!60}{$\triangleright$ Reuse top-$k$ indices from layer $l-1$}
\STATE $\mathcal{I}_l \gets \mathcal{I}_{l-1}$
\STATE $K^{sparse} = K^{cache}[\mathcal{I}_l]$
\STATE $V^{sparse} = V^{cache}[\mathcal{I}_l]$
\STATE \textcolor{blue!60}{$\triangleright$ Calculate sparse attention output}
\STATE $O \leftarrow \operatorname{Attention}(Q,K^{sparse},V^{sparse})$
\ENDIF
\ENDFOR
\end{algorithmic}
\end{algorithm}

Given a layer-wise execution policy $\pi = \{a_1, a_2, \dots, a_L\}$ obtained by dynamic programming through offline profiling, HyLRA applies a hybrid attention strategy during decoding. Instead of uniformly applying full attention at every layer, HyLRA selectively alternates between full attention and top-$k$ reuse attention, according to the policy $\pi$ for each layer.

For each newly generated token, decoding proceeds sequentially from the first transformer layer to the final layer. At each layer $l$, one of the following two execution modes is applied.

\textbf{Full attention at sensitive layers.}  
If layer $l$ is marked as a sensitive layer ($a_l = \text{Full}$), HyLRA performs standard full attention to preserve generation accuracy. In addition to computing the attention output, the top-$k$ attention indices $\mathcal{I}_l$ are extracted from the attention scores. These indices summarize the most relevant context positions at layer $l$ and are passed to the subsequent layer as candidate attention locations. Since full attention is already required at sensitive layers, extracting $\mathcal{I}_l$ introduces negligible additional overhead.

\textbf{Top-$k$ reuse attention at tolerant layers.}  
If layer $l$ is designated as a tolerant layer ($a_l = \text{Reuse}$), HyLRA reuses the top-$k$ indices $\mathcal{I}_{l-1}$ produced by the preceding sensitive layer. Attention computation at layer $l$ is then restricted to the corresponding subset of key-value pairs, avoiding a full scan over the entire KV cache. By operating on a reduced set of context positions, tolerant layers significantly reduce memory access and computation cost during decoding.

\textbf{Hybrid Decoding Strategy.}  
By interleaving sensitive and tolerant layers according to the pre-determined policy $\pi$, HyLRA periodically refreshes attention indices through full attention while maximizing efficiency via index reuse. This hybrid strategy exploits the strong inter-layer similarity of attention patterns during decoding, enabling substantial reductions in decoding latency with negligible impact on generation accuracy.

\subsection{Hardware-Efficient Optimizations}

To bridge the gap between theoretical FLOPs reduction and practical wall-clock latency improvements, we introduce two complementary system-level optimizations that target memory access irregularity and bandwidth limitations.

{\bf Adaptation to block-level sparsity.} While token-level sparsity enables fine-grained control over attention computation, it often leads to irregular and non-coalesced memory accesses, which can underutilize GPU memory bandwidth, particularly during decoding. HyLRA is agnostic to the granularity at which sparsity is applied and can be naturally adapted to block-level sparse attention mechanisms. In this setting, consecutive tokens are grouped into fixed-size blocks (e.g., block size $B=128$), and top-$k$ selection is performed at the block level rather than on individual tokens. For tolerant layers, HyLRA reuses the block indices selected in the preceding layer, leveraging the strong inter-layer stability of attention patterns observed during decoding. This coarse-grained reuse leads to more regular and contiguous memory accesses and allows HyLRA to integrate seamlessly with existing block-level sparse attention implementations, such as QUEST, without imposing additional algorithmic constraints.

{\bf Bandwidth reduction via index-guided offloading.} In long-context decoding scenarios where the KV cache exceeds on-device memory capacity, KV tensors are commonly offloaded to host memory, making PCIe bandwidth a critical performance bottleneck. HyLRA mitigates this overhead through index-guided offloading. By reusing the top-$k$ block indices identified in the preceding layer, HyLRA determines the subset of KV blocks required by the current tolerant layer prior to execution. This enables selective prefetching of only the critical KV blocks from CPU memory to GPU memory, instead of transferring the entire KV cache. By filtering out non-essential blocks on the host side and transferring only a minimal working set, HyLRA reduces redundant data movement over the PCIe bus and improves end-to-end inference latency in offloading-intensive decoding settings.

\section{Experiments}
\subsection{Setting}

{\bf Datasets.} We evaluate HyLRA on LongBench v2 \citep{bai2023longbench}, a comprehensive benchmark designed to assess the long-context capabilities of Large Language Models (LLMs). Our evaluation covers six diverse task categories provided by the benchmark: Single-document QA \citep{dasigi2021dataset}, Multi-document QA \citep{ho20202wikimultihopqa}, Long-dialogue History Understanding \citep{zhong2021qmsum}, Long Structured Data Understanding \citep{pasupat2015compositional}, Long In-context Learning \citep{cobbe2021training}, and Code Repository Understanding \citep{liu2023repobench}. These tasks encompass a wide range of input lengths and complexity levels, requiring the model to perform retrieval, reasoning, and synthesis over extensive contexts.

{\bf Models.} We select two models for our evaluation: DeepSeek-R1 \citep{deepseek2025} and QWQ-32B-W8A8 \citep{qwen2024}. DeepSeek-R1 represents a high-performance reasoning-oriented model, while QWQ-32B-W8A8 is a quantized version (8-bit weights and activations) of the Qwen reasoning model, allowing us to assess performance under memory-constrained settings. All evaluations were conducted on a high-performance compute node equipped with eight accelerator devices, providing an aggregate 768GB of HBM3 memory (96GB per device).

{\bf Baselines.} We compare HyLRA against two key baselines: Full Attention, which serves as a standard attention mechanism, and Jump3,  which utilizes a static variant of HyLRA that employs a fixed jump step of 3 instead of our adaptive mechanism. 

\subsection{Longbench Experiment}

\begin{table}[!t]
    \centering
    \caption{Performance comparison on LongBench v2 evaluating tasks with various difficulties (Easy, Hard) and context lengths (Short, Medium, Long) using DeepSeek-R1 and QWQ-32B-W8A8. HyLRA outperforms Jump3 at a 2048 token budget and achieves a superior overall performance than full attention. The best performance for each task is highlighted in bold.}
    \label{tab00}
    \resizebox{\linewidth}{!}{ 
    \begin{tabular}{llcccccc}
        \toprule
        \textbf{Model} & \textbf{Method} & \textbf{Overall} & \textbf{Easy} & \textbf{Hard} & \textbf{Short} & \textbf{Medium} & \textbf{Long} \\
        \midrule
        \multirow{3}{*}{DeepSeek-R1} 
        & Full & 44.33 & 48.44 & 41.80 & 48.89 & 40.00 & \textbf{45.37} \\
        & Jump3 & 42.54 & 43.75 & 41.80 & 50.00 & 39.07 & 37.04 \\
        & HyLRA & \textbf{46.32} & \textbf{49.48} & \textbf{44.37} & \textbf{52.78} & \textbf{43.72} & 40.74 \\
        \midrule
        \multirow{3}{*}{QWQ-32B-W8A8} 
        & Full & \textbf{45.73} & \textbf{48.96} & \textbf{43.73} & \textbf{52.22} & \textbf{44.19} & 37.96 \\
        & Jump3 & 40.76 & 43.75 & 38.91 & 45.56 & 38.14 & 37.96 \\
        & HyLRA & 45.03 & \textbf{48.96} & 42.61 & 50.00 & 42.79 & \textbf{41.20} \\
        \bottomrule
    \end{tabular}
    }
\end{table}

\begin{table*}[!t]
    \centering
    \caption{Detailed performance on LongBench v2. We present the specific scores for each sub-task and the average scores for each major category (e.g., Single/Multi-document QA, Code Understanding) across DeepSeek-R1 and QWQ-32B-W8A8. \textbf{Bold} indicates the best performance among the compared methods for each model.}
    \label{tab01}
    \small 
    {
    \begin{tabular}{lcccccc}
        \toprule
        & \multicolumn{3}{c}{\textbf{DeepSeek-R1}} & \multicolumn{3}{c}{\textbf{QWQ-32B-W8A8}} \\
        \cmidrule(lr){2-4} \cmidrule(lr){5-7}
        \textbf{Task / Dataset} & Full & Jump3 & HyLRA & Full & Jump3 & HyLRA \\
        \midrule
        
        \multicolumn{7}{l}{\textit{Single-document QA}} \\ 
        \textbf{Overall}                        & 42.29 & {\bf 48.00} & {\bf 48.00} & {\bf 48.00} & 47.43 & 46.86 \\
        Single-document QA: Academic            & 43.18 & {\bf 47.73} & 45.45 & 50.00 & {\bf 52.27} & 50.00 \\
        Single-document QA: Detective           & 22.73 & 22.73 & {\bf 31.82} & 31.82 & 36.36 & {\bf 45.45} \\
        Single-document QA: Event ordering      & {\bf 35.00} & {\bf 35.00} & 25.00 & {\bf 60.00} & 50.00 & 37.50 \\
        Single-document QA: Financial           & 54.55 & 63.64 & {\bf 68.18} & 50.00 & 50.00 & {\bf 52.27} \\
        Single-document QA: Governmental        & 33.33 & 33.33 & {\bf 44.44} & 38.89 & {\bf 55.56} & 36.11 \\
        Single-document QA: Legal               & 52.63 & {\bf 73.68} & {\bf 73.68} & 42.11 & 36.84 & {\bf 44.74} \\
        Single-document QA: Literary            & 50.00 & {\bf 56.67} & 50.00 & {\bf 56.67} & 46.67 & 53.34 \\
        \midrule
        
        \multicolumn{7}{l}{\textit{Multi-Document QA}} \\
        \textbf{Overall}                       & 44.80 & 40.00 & {\bf 49.60} & 44.80 & 43.20 & {\bf 45.20} \\
        Multi-document QA: Academic            & {\bf 62.00} & 44.00 & 56.00 & {\bf 62.00} & 56.00 & 55.00 \\
        Multi-document QA: Financial           & 46.67 & {\bf 53.33} & {\bf 53.33} & {\bf 46.67} & 33.33 & 40.00 \\
        Multi-document QA: Governmental        & 26.09 & 21.47 & {\bf 43.48} & 26.09 & {\bf 34.78} & {\bf 34.78} \\
        Multi-document QA: Legal               & 28.57 & {\bf 57.14} & {\bf 57.14} & 28.57 & 35.71 & {\bf 35.72} \\
        Multi-document QA: Multi-news          & {\bf 34.78} & 30.43 & {\bf 34.78} & 34.78 & 34.78 & {\bf 43.48} \\
        \midrule

        \multicolumn{7}{l}{\textit{Long-dialogue History Understanding}} \\
        \textbf{Overall}                       & 35.90 & 28.21 & {\bf 41.03} & 41.03 & 25.64 & {\bf 44.87} \\
        Long-dialogue History Understanding: Agent history QA            & 25.00 & 15.00 & {\bf 30.00} & {\bf 35.00} & 10.00 & 30.00 \\
        Long-dialogue History Understanding: Dialogue history QA         & 47.37 & 42.11 & {\bf 52.63} & 47.37 & 42.11 & {\bf 60.53} \\
        \midrule

        \multicolumn{7}{l}{\textit{Long Structured Data Understanding}} \\
        \textbf{Overall}                       & 24.24 & 15.15  & {\bf 27.27} & {\bf 24.24} & 18.18 & {\bf 24.24} \\
        Long Structured Data Understanding: Knowledge graph reasoning          & 33.33 & 13.33 & {\bf 40.00} & {\bf 33.33} & 13.33 & 26.67 \\
        Long Structured Data Understanding: Table QA          & {\bf 16.67} & {\bf 16.67} & {\bf 16.67} & 16.67 & 22.22 & {\bf 22.23} \\
        \midrule

        \multicolumn{7}{l}{\textit{Long In-context Learning}} \\
        \textbf{Overall}                       & 50.62 & 49.38  & {\bf 51.85} & {\bf 53.09} & 40.74 & 46.92 \\
        Long In-context Learning: Many-short learning          & 57.14 & {\bf 61.90} & {\bf 61.90} & {\bf 61.90} & 52.38 & 42.86 \\
        Long In-context Learning: New language translation          & {\bf 75.00} & 50.00 & 65.00 & 60.00 & 50.00 & {\bf 65.00} \\
        Long In-context Learning: User guide QA      & 35.00 & {\bf 42.50} & 40.00 & {\bf 45.00} & 30.00 & 40.00 \\
        \midrule

        \multicolumn{7}{l}{\textit{Code Repository Understanding}} \\
        \textbf{Overall}                       & {\bf 56.00} & 48.00  & 40.00 & 46.00 & 38.00 & {\bf 49.00} \\
        Code Repository Understanding: Code repo QA          & {\bf 56.00} & 48.00 & 40.00 & 46.00 & 38.00 & {\bf 49.00} \\
        \bottomrule
    \end{tabular}
    }
\vspace{15pt}
    \caption{Efficiency comparison across varying sequence length (from 8K to 60K). We present average time to first token (TTFT), generation/overall throughput, and speedup compared to full attention. Note that HyLRA achieves higher speedups as the sequence length increases.}
    \label{tab02}
    
    \small
    \renewcommand{\arraystretch}{1.2}
    
    \setlength{\tabcolsep}{4pt} 
    
    \resizebox{\textwidth}{!}{
    \begin{tabular}{c ccc ccc ccc cc}
        \toprule
        & \multicolumn{3}{c}{\textbf{AVG\_TTFT}} 
        & \multicolumn{3}{c}{\textbf{Generation Throughput}} 
        & \multicolumn{3}{c}{\textbf{Overall Throughput}} 
        & \multicolumn{2}{c}{\textbf{Speedup}} \\ 
        
        \cmidrule(lr){2-4} \cmidrule(lr){5-7} \cmidrule(lr){8-10} \cmidrule(lr){11-12}
        
        \textbf{Sequence Length} & Full & Jump3 & {\bf HyLRA} & Full & Jump3 & {\bf HyLRA} & Full & Jump3 & {\bf HyLRA} & Jump3 & {\bf HyLRA} \\
        \midrule
        
        8K & 
        16.81 & 15.27 & {\bf 14.85} & 
        1225 & {\bf 1229} & 1145 & 
        11033 & {\bf 11068} & 10309 &
        1.00 & 0.93 \\ 
        
        16K & 
        26.55 & {\bf 25.69} & 26.23 & 
        728 & {\bf 839} & 771 & 
        6555 & {\bf 7553} & 6946 &
        1.15 & 1.06 \\ 
        
        30K & 
        38.23 & {\bf 37.25} & 37.76 & 
        450 & {\bf 544} & 514 & 
        3826 & {\bf 4627} & 4374 &
        1.21 & 1.14 \\ 
        
        60K & 
        57.69 & {\bf 53.98} & 56.62 & 
        177 & {\bf 273} & 258 & 
        2835 & {\bf 4730} & 4128 &
        1.54 & 1.45 \\ 
        
        \bottomrule
    \end{tabular}
    }
\end{table*}

We conduct a comprehensive evaluation of HyLRA on LongBench v2 to assess its capabilities across diverse real-world long-context scenarios. As the results in Table~\ref{tab00} demonstrate, HyLRA exhibits superior efficiency compared to baselines. When strictly constrained to the same 2,048-token budget, HyLRA consistently outperforms Jump3 across all tested models. Remarkably, HyLRA not only surpasses the sparse baseline but also achieves a competitive score compared to standard full attention. This performance advantage proves to be robust, showing consistent gains across varying levels of task difficulty and spanning the full spectrum of context lengths.

The detailed analysis provided in Table~\ref{tab01} further underscores the versatility of our approach, showing that HyLRA attains either the best or highly competitive performance across major distinct task categories. Collectively, these results confirm that HyLRA successfully preserves critical contextual information while filtering out irrelevant noise. By significantly reducing attention computation without compromising accuracy, HyLRA establishes itself as a highly effective and robust solution for efficient long-context inference.

\subsection{Efficiency Analysis}

We evaluate the efficiency of HyLRA under increasing sequence lengths ranging from 8K to 60K. Table~\ref{tab02} shows time to first token (TTFT), generation throughput, overall throughput, and the corresponding speedup over full attention.

Across all sequence lengths, HyLRA achieves competitive TTFT and throughput compared to Jump3, while full attention remains slightly faster at 8K tokens. As the sequence length increases, HyLRA exhibits a steadily increasing speedup, demonstrating strong scalability in long-context inference. At 60K tokens, HyLRA achieves up to 1.45× overall speedup, confirming its effectiveness in reducing attention computation while maintaining efficient generation.

\section{Conclusion}
We present HyLRA, an efficient hybrid layer-wise attention framework that exploits heterogeneity in layer sensitivity to reduce long-context inference overhead. HyLRA dynamically combines full attention on sensitive layers with index reuse on tolerant layers, significantly reducing quadratic attention computation and KV-cache processing without sacrificing accuracy. Comprehensive evaluations show that HyLRA achieves up to 1.45$\times$ overall inference speedup with negligible accuracy loss ($< 1\%$), consistently outperforming baselines. These results demonstrate that layer-aware hybrid attention is a practical and scalable approach for efficient long-context LLM inference.

\bibliography{example_paper}
\bibliographystyle{icml2026}




\end{document}